\documentclass[journal]{IEEEtran}
\makeatletter
\def\ps@headings{%
\def\@oddhead{\mbox{}\scriptsize\rightmark \hfil \thepage}%
\def\@evenhead{\scriptsize\thepage \hfil \leftmark\mbox{}}%
\def\@oddfoot{}%
\def\@evenfoot{}}
\makeatother

\usepackage{graphics}
\usepackage{amsmath,graphicx}
\usepackage[lofdepth,lotdepth]{subfig}
\usepackage{paralist,cite}
\usepackage{amssymb,color}
\usepackage{algorithm2e}
\usepackage{algorithmic}

\newcommand{\nix}[1]{}

\begin{document}
\title{Pilgrims Face  Recognition  Dataset -- HUFRD}
\author{
Salah A. Aly\\
\medskip
Center of Research Excellence in Hajj and Umrah  (HajjCore), \\ College of Computers and Information Systems, Umm Al-Qura University, Makkah, KSA\\ Email:  salahaly@uqu.edu.sa
}
\maketitle

\begin{abstract}
In this  work, we define a new pilgrims face recognition and face detection dataset, called Hajj and Umrah facial dataset. The new developed dataset presents various pilgrims' images taken from outside the Holy Masjid El-Harram in Makkah during the 2011-2012 Hajj  and Umrah seasons. Such dataset will be used to test our developed facial recognition and detection algorithms, as well as in the missing and found recognition system~\cite{crowdsensing}.

\end{abstract}

\section{Introduction}
\label{sec:intro}

\begin{figure}[h]
\centering
    \includegraphics[width=8.5cm,height=14.5cm]{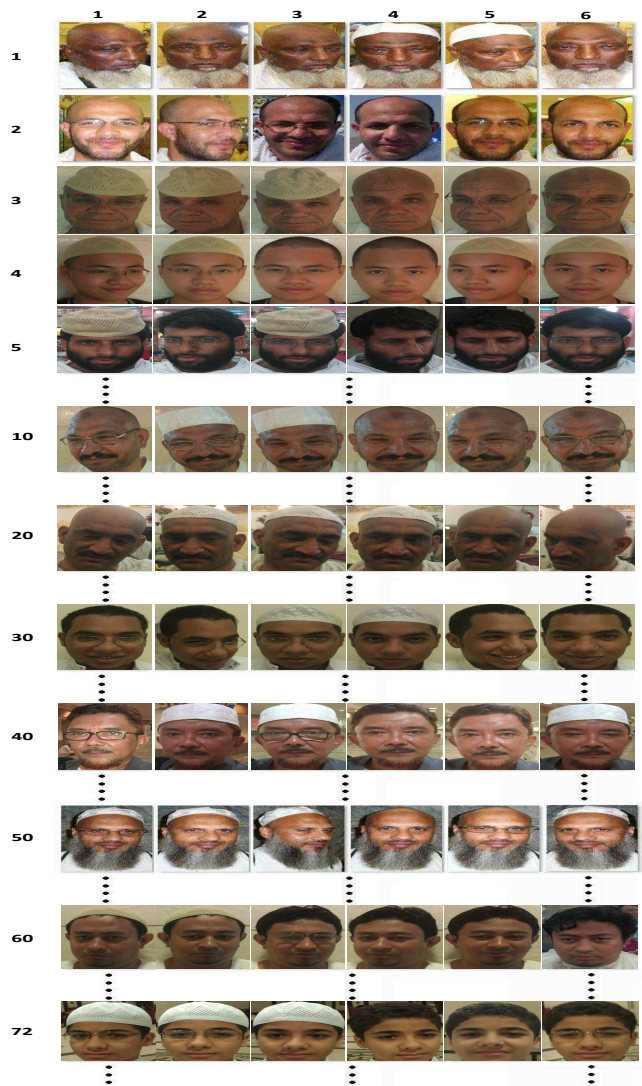}
    \caption{The new Hajj and Umrah facial recognition dataset presents various images taken outside El-Harram in Makkah~\cite{crowdsensing} 2011-2012 Hajj  and Umrah seasons.}
    \label{fig:hajjds}
\end{figure}

Every year millions of muslims arrive to perform the holy rituals of Hajj and Umarh in Makkah Al-Mokarramah, Kingdom of Saudi Arabia. During these rituals the Saudi authorities save no effort to facilitate the stay of the pilgrims in the kingdom. The greatest challenge is the huge number of missing persons and unidentified deaths every year. So, an efficient monitoring system is essential to provide means of tracking missing and found individuals and identifying them in order to take the necessary actions. To solve this problem, we describe a new pilgrims face recognition dataset, HUFRD, which will be used in the CrowdSensing system for recognizing missing and found people~\cite{crowdsensing}, see sample classes of HUFRD in Fig.~\ref{fig:hajjds}.

Facail recognition   is one of the most complex applications in the field of Computer Vision. Developing a computational model of human faces for detection and recognition, though interesting, can prove to be a very challenging task. The rapid  progress in PCs' speed certainly assists the  operation of face detection and recognition to be done in real-time.

One of the most crucial steps in the CrowdSensing system that we describe in~\cite{crowdsensing}  is to how to  detect the individual faces in an image and run a  face recognition algorithm through a database of registered pilgrims. Therefore, the proposed approach is more focused on developing a pattern detection algorithm that does not depend on the three-dimensional complex data of the face, but it depends on the general outer Silhouette that is almost shared between all faces. Our goal is to develop a computational model for face detection and recognition that is reasonably simple, and acceptably accurate under various conditions of lighting, facial expressions, and background environment.

\begin{figure}[t]
\centering
    \includegraphics[width=7cm,height=2.5cm]{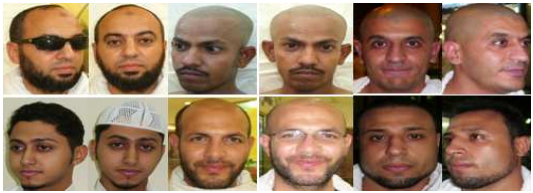}
    \caption{A sample set of Hajj and Umrah Facial Recognition Dataset (1)}
    \label{fig:fig1}
\end{figure}

In general, facial dataset type has a strong effect in face recognition performance, it contains multiple images for each person. Facial datasets depend on several different factors such as:
\begin{compactitem}
\item  Facial expression  such as sadness, happiness,  and  facial pose.
\item  Occlusion: faces may be partially occluded by other objects (like wearing glasses ).
\item  Imaging conditions like lighting and camera resolution.
\item  Presence or absence of structural components like  beards, mustaches and glasses.
\end{compactitem}


\section{Pilgrims Face Recognition Dataset}
\label{sec:systemdescription}

In the Hajj and Umrah face recognition dataset, HUFRD, hundreds of pilgrims images are taken randomly during 2011-2012 Hajj and Umrah seasons. The dataset contains faces from more than $25$ countries as shown in Fig.~\ref{fig:fig1}. For each person, 6 up to 15 images are taken to define a person's class in the dataset. Further information about this dataset can be found in CrowdSensing project~\cite{crowdsensing}.

\subsection{Sample Images of Hajj and Umrah Face Recognition Dataset}

\begin{figure}[h]
\centering
    \includegraphics[width=0.4\textwidth]{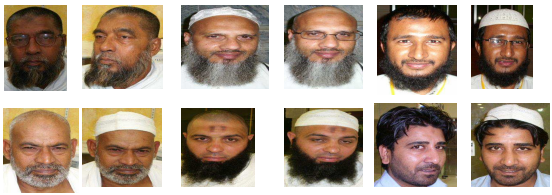}
    \caption{A sample set of Hajj and Umrah Facial Recognition Dataset  (2)}
    \label{fig:fig2}
\end{figure}

The HUFRD contains images taken during the 2011-2012 Hajj and  Umrah seasons of a large number of pilgrims (varied races and appearances). It contains at least six images for each individual, in a varied range of poses, facial expressions (open and closed eyes, smiling and not smiling) and facial details (glasses/ no glasses), and in  different lighting conditions and against random backgrounds.
All images are in full-color JPG format, see Fig.~\ref{fig:fig1} and`\ref{fig:fig2}.

\subsection{CrowdSensing System}

The CrowdSensing system is established to support
the existing efforts to manage the crowds and solve the missing
and found problem during Hajj and Umrah seasons in KSA.
The goal of this CrowdSensing system is to use techniques
from Computer Vision and Image Processing to develop a portal
website for Hajj and Umrah missing and found people~\cite{crowdsensing}. The system
requires all pilgrims to register their personal data when they
plan to perform Hajj or Umrah.

\begin{figure}[t]
  \begin{center}
  \includegraphics[width=8.3cm,height=6cm]{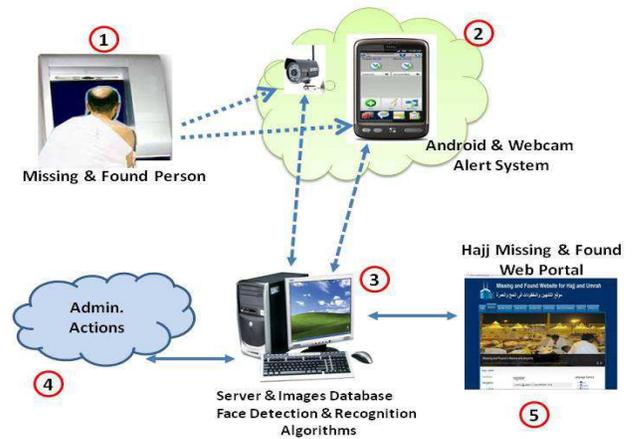}
  \caption{MFHajj portal interface website developed by the Crowdsensing.net team. The system consists of data collectors (mobiles, cameras, PCs), main server, search engine, and an alerting system. The system is used to recognize missing and found people during Hajj and Umrah seasons}\label{fig:hajjmfsys}
  \end{center}
\end{figure}

One application of the proposed dataset is the CrowdSensing system, which consists of three main components, see Fig.~\ref{fig:hajjmfsys}:
 \begin{enumerate}
   \item A database of all individuals arriving at the kingdom to perform the holy rituals. This database contains all their personal information along with a personal photo, and it can be updated via our web portal.
   \item  Advanced monitoring cameras scattered around the Grand Mosque in Makkah, airports, hospitals, and all areas of interest.
   \item  Our proposed face detection \& recognition algorithm is to be used for acquiring faces from images captured by the monitoring cameras and use them to identify missing and found individuals ~\cite{crowdsensing}.
 \end{enumerate}

\section{Other Face Detection and Recognition Datasets}
\label{sec:related}

Some other face recognition research work can be found in~\cite{oneFace,jyoti2011,soft2005,belongie2002,Yang04}, and face recognition datasets\cite{JAFFE,UMIST,ORL,Haar01}. We briefly describe a note about each known face recognition datataset such as FAFFE, UMIST, ORL, and FERET.

\subsection{ORL Dataset}

\begin{figure}[h]
  \begin{center}
  \includegraphics[width=7cm,height=4cm]{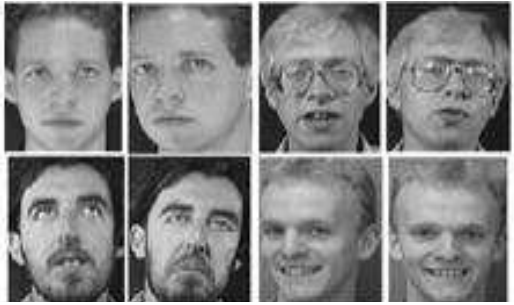}
  \caption{Samples  of facial images: ORL Dataset}\label{fig:orl}
  \end{center}
\end{figure}

The AT\&T Database of Faces, (formerly 'The ORL Database of Faces'), see~\ref{fig:orl}. The database was used in the context of a face recognition project carried out in collaboration with the Speech, Vision and Robotics Group of  Engineering Department at Cambridge University.
There are ten different images of each of $40$ distinct subjects. For some subjects, the images were taken at different times, varying the lighting and facial expressions. All images were taken against a dark homogeneous background with the subjects in an upright, frontal position (with tolerance for some side movement). The images are in PGM format. The size of each image is $92x112$ pixels, with $256$ grey levels per pixel~\cite{ORL}.

\subsection{Japanese Dataset}

\begin{figure}[h]
  \begin{center}
  \includegraphics[width=7cm,height=4cm]{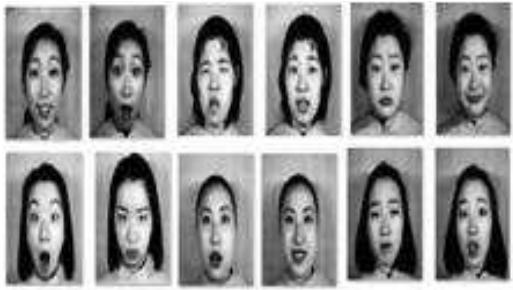}
  \caption{Samples  of facial images: Japanese Dataset }\label{fig:Japanese}
  \end{center}
\end{figure}

The Japanese database consists of $10$ persons. The images per person vary from $20$ to $23$, where facial expressions are varying~\cite{JAFFE}, see Fig.~\ref{fig:Japanese}.
\bigskip

\subsection{UMIST Dataset}

\begin{figure}[h]
  \begin{center}
  \includegraphics[width=7cm,height=4cm]{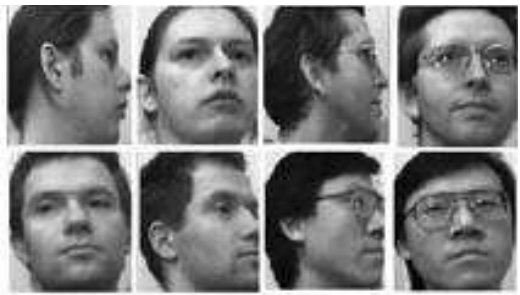}
  \caption{Samples  of facial images: UMIST dataset}\label{fig:umist}
  \end{center}
\end{figure}

The Sheffield (previously UMIST) face database consists of $564$ images of $20$ individuals (mixed race/gender/appearance).  Images are numbered consecutively as they were taken. The images are all in PGM format, approximately $220 x 220$ pixels with $256$-bit grey-scale~\cite{UMIST}, see Fig.~\ref{fig:umist}.

\subsection{FERET}
\begin{figure}[h]
  \begin{center}
  \includegraphics[width=7cm,height=4cm]{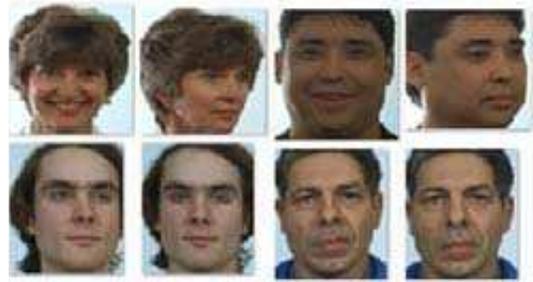}
  \caption{Samples  of facial images: FERET dataset}\label{fig:feret}
  \end{center}
\end{figure}

The  Face Recognition Technology (FERET)  dataset is one of the known face recognition datasets that was collected between 1993 and 1996 and was developed by DARPA~\cite{Haar01}, see Fig.~\ref{fig:feret}.

\subsection{The Yale Face Dataset}

Yale face dataset images should be in grayscale and GIF format~\cite{yale}. There are 11 different images of each of 15 distinct subjects, these different images  mainly depend on different face expressions and little effect of light. For the same subject there are three different images which light is focused in different directions such as  center, Lift, Right, and the other 8 images related with person expression like happy, sad, sleepy, Shocked, with or no glasses.

\begin{figure}[h]
\centering
            \includegraphics[width=7cm,height=2cm]{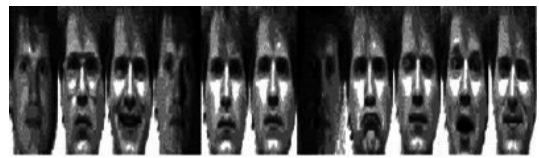}
            \caption{Samples of facial images: Yale face dataset}
            \label{HajjRes}
\end{figure}

\subsection{PIE Dataset}

PIE face dataset  images in PIE Dataset are colored, it mainly depend on few face expressions, face pose and illumination condition with only one record session. There are 68 distinct subjects and for each subject, there are 13 different poses, 43 different illumination with 4 different expression.~\cite{PIE}.

\begin{figure}[h]
\centering
            \includegraphics[width=7cm,height=4cm]{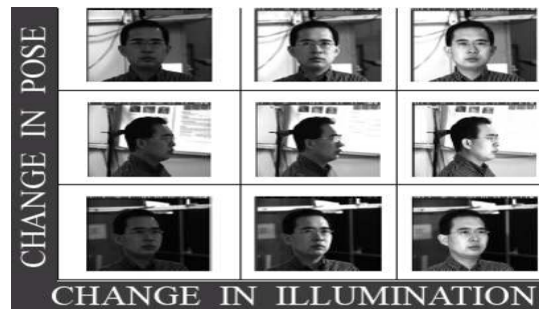}
            \caption{Samples of facial images: PIE dataset}
            \label{HajjRes}
\end{figure}

\subsection{Multi-PIE Dataset}

Multi-PIE  face dataset  images in Multi-PIE Dataset are colored, and mainly depend on face expressions, face pose and illumination condition with multiple record session. There are 377 distinct subjects and for each subject there are 15 different poses, 19 different illumination with 6 different expression with 4 sessions.
\begin{figure}[h]
\centering
            \includegraphics[width=8cm,height=4cm]{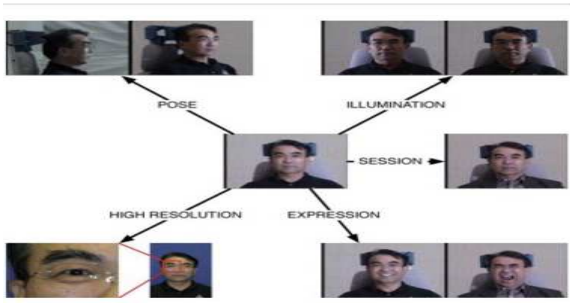}
            \caption{Samples of facial images: Multi-PIE dataset}
            \label{HajjRes}
\end{figure}
In both PIE and Multi-PIE,  they support a 3D Room with 13 camera.  Also, for illumination condition,  they support a 3D room with a flash system. Using multiple cameras which are fixed in specific places in the 3D room save time and efforts for preparing my dataset specially for large dataset
as each person in dataset only sits in a chair with and fix his head.

\subsection{AR Face Dataset}

AR face dataset images in AR face Dataset are colored, and depend on face expressions such as neutral, smile, anger, scream,  also, illumination condition (left, right, center)  and occlusions (sun glasses and scarf) with two record sessions during two weeks. There are 126 distinct subjects (over 4,000 color images) divided between man and women (70 men and 56 women)~\cite{AR}.
\begin{figure}[h]
\centering
            \includegraphics[width=8cm,height=3cm]{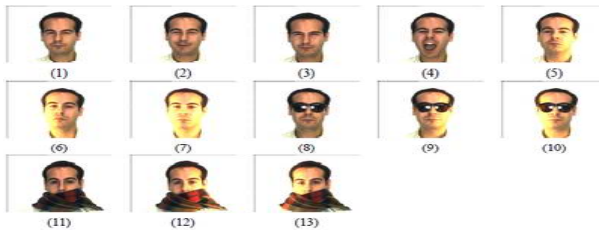}
            \caption{Samples of facial images: AR face dataset}
            \label{HajjRes}
\end{figure}

\section{Discussion and Conclusion}

In this note, we have described a new Hajj and Umrah face recognition dataset, HUFRD, for face recognition and face detection. During  the huge overcrowds that occur in the two holy cities of Makkah and Madina in KSA every year (more than $5$ millions people observe Hajj for at least $10$ days), in addition to five millions people observe Umrah throughout the year. The proposed CrowdSensing system can be used to identify missing, found, and dead parsons.  Further details about this system can be found in the project website~\cite{crowdsensing}.

\begin{table}[h]\begin{center}
\caption{Information about  ORL, UMIIST,  Japanese, FERET, HUFRD datasets}
\begin{tabular}{|c|c|c|}
\hline
\hline
&&\\
dataset& \# persons & \# images\\
&& \\
\hline
ORL &	40 persons,& Each  has 10 images\\
	UMIST	& 20 persons, & vary from 19 to 48\\
	JAFFE	&10 persons, & vary from 20 to 23\\
FERET& too many& many \\
HUFRD&too many&each has at least 6 to 15\\
Yale face&15 persons&each has 11 images\\
PIE&68 persons&each has 60 images  \\
Multi-PIE&337 persons&each has 40 images for one session\\
AR face&126persons& each has 26 images \\
&&\\
\hline
\end{tabular}\end{center}
\end{table}
\bigskip

\section*{Acknowledgments}

This work is funded by a  grant number 11-INF1707-10 from the Long-Term National Plan for Science, Technology and Innovation (LT-NPSTI), the King Abdulaziz City for Science and Technology (KACST), Kingdom of Saudi Arabia. We thank the Science and Technology Unit at Umm A-Qura University for their continued logistics support.

S.~A.~A. would like to thank Hossam Zawbaa, Dr. Motaz Abdelwahab, Dr. Andan A. Gutub, and all staff members at HajjCore for their comments and feedbacks of earlier draft of this work.

S.~A.~A. is thankful and grateful to all pilgrims who kindly participated in this study. We thank Allah for his mercy and ultimate support.


\onecolumn
\begin{figure}
\centering
    \includegraphics[width=17cm,height=21cm]{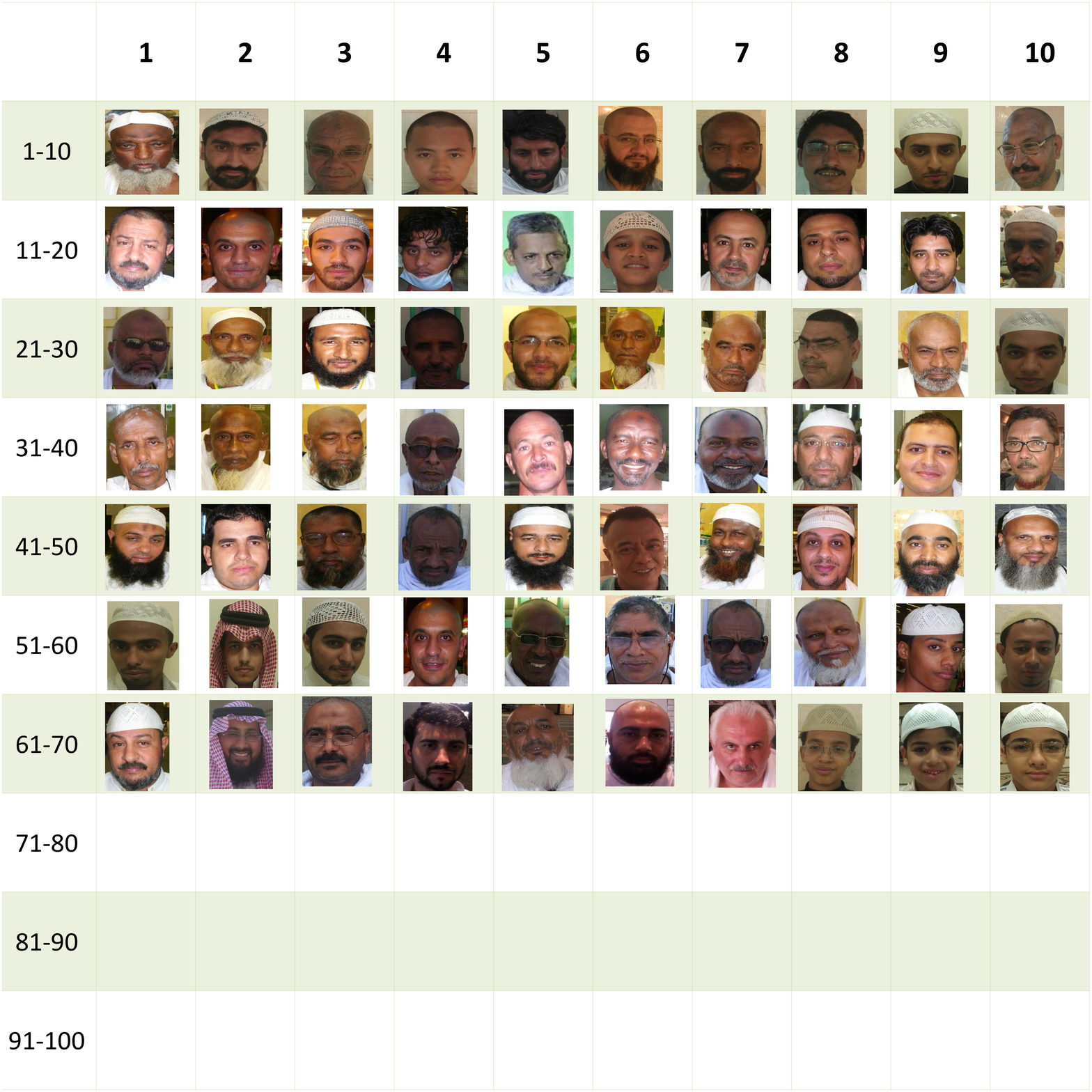}
    \caption{Pilgrims facial  dataset (HUFRD) for face detection and recognition}
    \label{fig:hajjds2}
\end{figure}
\end{document}